\documentclass[10pt,twocolumn,letterpaper]{article}
\usepackage[accsupp]{axessibility}  
\usepackage{iccv}
\usepackage{times}
\usepackage{epsfig}
\usepackage{graphicx}
\usepackage{amsmath}
\usepackage{amssymb}
\usepackage{booktabs}
\usepackage{multirow}
\usepackage{algorithm}
\usepackage{algorithmic}
\usepackage{hhline}
\usepackage{color}
\usepackage{listings}
\usepackage{colortbl}
\usepackage{threeparttable}
\usepackage{pifont}
\usepackage{footnote}
\usepackage{marvosym}
\usepackage{footmisc}

\usepackage[pagebackref=true,breaklinks=true,colorlinks,bookmarks=false]{hyperref}

\usepackage[capitalize]{cleveref}
\crefname{section}{Sec.}{Secs.}
\Crefname{section}{Section}{Sections}
\Crefname{table}{Table}{Tables}
\crefname{table}{Tab.}{Tabs.}

\def\etal{{\it{et al.}}}
\def\ie{{\it{i.e.}}}

\iccvfinalcopy 

\ificcvfinal\fi

\begin{document}

\title{Towards General Low-Light Raw Noise Synthesis and Modeling}

\author{
   \hspace{-0.4cm} Feng Zhang$^{1}$
     \hspace{0.01cm} 
    Bin Xu$^{2}$
     \hspace{0.01cm} 
    Zhiqiang Li$^{1,2}$
     \hspace{0.01cm} 
    Xinran Liu$^{1}$ 
     \hspace{0.01cm}
    Qingbo Lu$^{2}$ 
     \hspace{0.01cm}
    Changxin Gao$^{1}$
    \hspace{0.01cm}
    Nong Sang$^{1\dag}$\\
$^1$National Key Laboratory of Multispectral Information Intelligent Processing Technology,\\ School of Artificial Intelligence and Automation, Huazhong University of Science and Technology\\
$^2$DJI Technology Co., Ltd \\
{\tt\small\{fengzhangaia, lxryx, cgao, nsang\}@hust.edu.cn,
\{mila.xu, cristopher.li, qingbo.lu\}@dji.com}
}

\maketitle
\let\thefootnote\relax\footnotetext{$\dag$ Corresponding authors.}
\ificcvfinal\thispagestyle{empty}\fi

\begin{abstract}
Modeling and synthesizing low-light raw noise is a fundamental problem for computational photography and image processing applications. Although most recent works have adopted physics-based models to synthesize noise, the signal-independent noise in low-light conditions is far more complicated and varies dramatically across camera sensors, which is beyond the description of these models. To address this issue, we introduce a new perspective to synthesize the signal-independent noise by a generative model. Specifically, we synthesize the signal-dependent and signal-independent noise in a physics- and learning-based manner, respectively. In this way, our method can be considered as a general model, that is, it can simultaneously learn different noise characteristics for different ISO levels and generalize to various sensors. Subsequently, we present an effective multi-scale discriminator termed Fourier transformer discriminator (FTD) to distinguish the noise distribution accurately. Additionally, we collect a new low-light raw denoising (LRD) dataset for training and benchmarking. Qualitative validation shows that the noise generated by our proposed noise model can be highly similar to the real noise in terms of distribution. Furthermore, extensive denoising experiments demonstrate that our method performs favorably against state-of-the-art methods on different sensors.
\end{abstract}

\begin{figure}[t]
\centering
\includegraphics[width=\linewidth]{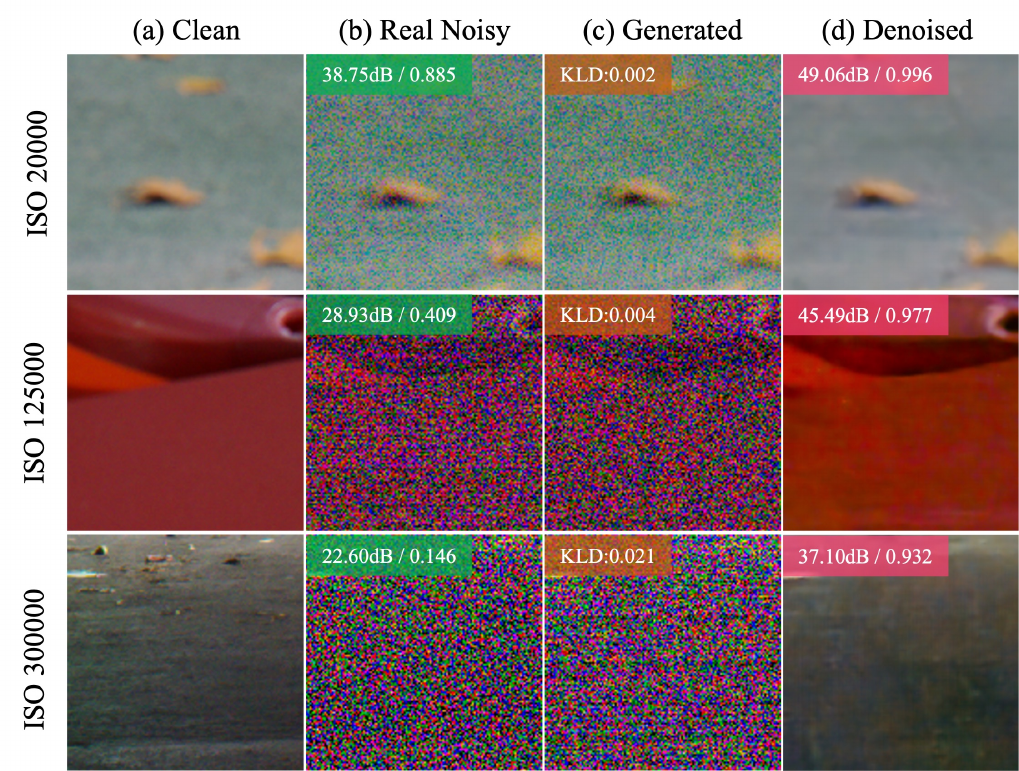}
\caption{Examples of generated and denoised raw images from our proposed method. (a) Clean raw image, (b) Real noisy raw image, (c) Generated noisy raw image from our proposed noise model, (d) Denoised raw image of the denoise network~\cite{chen2018learning} trained on the image generated by our noise model. Our model can generate noise at different ISO levels by a single learned noise model. The results of the visualization and quantitative metric Kullback-Leibler divergence (KLD) demonstrate that the distribution of the synthetic noise is close to the real noise distribution. Noise at high ISO levels drowns out the signal, resulting in unsatisfactory image denoising.}
\label{fig:intro}
\end{figure}

\section{Introduction}
\label{intro}
Low-light raw denoising is an important yet challenging problem for the increasingly widespread computational photography~\cite{wang2020practical}. Due to the powerful computational capability of deep learning, the learning-based low-light raw denoising algorithms have shown great superiority and become the mainstream manner~\cite{chen2018learning,feng2022learnability} in recent years. However, since the standard paradigm of deep learning is to learn a mapping from the low-light noisy raw image to its normal-light counterpart, this leads to a reliance on the large-scale dataset of real-world noisy-clean raw image pairs, which is extremely tedious and labor-intensive to collect.

A naive strategy is to synthesize the low-light noisy raw images to obtain more paired training data. Existing noise synthesis methods can be roughly categorized into physics-based noise models and Deep Neural Network (DNN)-based noise models. 

Physics-based noise modeling~\cite{foi2008practical, wei2020physics, wang2019enhancing} is the most commonly used noise synthesis method in low-light conditions, which obtains the statistical distribution of different noise sources by analyzing the physical process of camera sensor imaging. However, noise sources on different camera sensors vary widely due to differences in circuit design and signal processing techniques, making it impossible for physics-based methods to extract and model all noise sources accurately. Moreover, each noise source's properties and statistical behavior vary significantly at different exposure times or ISO levels, making physics-based methods tedious and error-prone. All these limitations make it impossible for a physics-based method to achieve accurate noise modeling on multiple camera sensors.

DNN-based noise modeling~\cite{abdelhamed2019noise, chang2020learning} learns to synthesize noise from real captured datasets with deep generative networks. Although the existing deep models show promising synthetic results on raw images due to their powerful representation capability, some previous studies~\cite{zhang2021rethinking,monakhova2022dancing} have revealed that they perform poorly on extremely low-light raw images.

In this paper, we present a new perspective on synthesizing realistic low-light raw noise. Specifically, instead of directly synthesizing noise with generative networks, we separate the noise synthesis process into two components, \ie, signal-dependent and signal-independent, which are implemented through a physics-based manner and a learning-based manner, respectively. We employ a pre-trained denoise network during the training procedure to transfer the synthesized and real noisy raw images into a nearly noise-free image space to perform image domain alignment. Meanwhile, to better distinguish the synthesized and real noise, we present an effective multi-scale discriminator, namely Fourier transformer discriminator (FTD), to perform the noise domain alignment. In addition, we collect a new low-light raw denoising (LRD) dataset for training and benchmarking. Extensive experiments demonstrate that our noise model performs favorably against existing state-of-the-art noise models on different camera sensors. Fig.~\ref{fig:intro} shows examples of the synthetic noisy raw images and the corresponding denoising results trained on the synthetic noisy raw images. 

In conclusion, our contributions can be summarized into three aspects:
\vspace{-0.5em}
\begin{itemize}
\vspace{-0.5em}
\item We propose a general noise model with separated synthesis processes to express the noise terms of according characteristics, enabling the noise model to imitate accurate low-light raw noise on different sensors.
\vspace{-0.5em}
\item We establish an effective multi-scale discriminator framework, namely Fourier transformer discriminator (FTD), which encourages the generator to favor solutions that reside on the manifold of real low-light raw noise distributions.
\vspace{-0.5em}
\item We collect a new large-scale dataset for low-light raw denoising benchmarking and researching.
\end{itemize}

\section{Related Work}
\label{related}
\textbf{Deep Image Denoising.} Image denoising is an extensively-studied yet still unresolved issue in computational photography. In the designing of traditional denoising algorithms, making an analytical regularizer related to image priors (e.g., sparsity~\cite{elad2006image, aharon2006k}, smoothness~\cite{portilla2003image, rudin1992nonlinear}, self-similarity~\cite{buades2005non, dabov2007image, maggioni2012video}, low rank~\cite{gu2014weighted}) plays a critical role. In the modern era, deep learning-based algorithms~\cite{zhang2017beyond, brooks2019unprocessing, guo2019toward, yue2019variational} have demonstrated their powerful superiority. However, most of them are based on the assumption that the noise conforms to a Gaussian distribution, but the noise captured in the real world is much more complex than Gaussian noise, which makes these methods even inferior to the traditional method BM3D~\cite{dabov2006image}. To address this problem, several works have established a database of noisy and clean image pairs taken by real cameras as a benchmark~\cite{abdelhamed2018high, plotz2017benchmarking}, thus improving the denoising performance of learning-based algorithms~\cite{chen2015learning, chen2018deep, gharbi2016deep} in real-world scenes. Although this line of work is promising, the burden of acquiring real image pairs is heavy, and the collected pairs suffer from pixel misalignment, luminance misalignment, limited data volume, and lack of diversity.

\textbf{Physics-based Noise Model.} The additive white Gaussian (AWGN) noise model is the most widely-used physics-based noise model. However, it deviates strongly from the realistic noise distribution, which leads to significant performance degradation on images with real noise~\cite{plotz2017benchmarking,abdelhamed2018high}. The classical Poisson-Gaussian (P-G) noise model~\cite{foi2008practical,foi2009clipped, hasinoff2014photon, brooks2019unprocessing, mildenhall2018burst, wang2019enhancing} is proposed to handle the domain gap between synthetic images and real images, which considers both photon shot noise and sensor readout noise. Most recently, Wei~\etal~\cite{wei2020physics} has developed a novel noise model by analyzing the noise generation process in the image processing pipeline and obtaining the distribution of noise sources by using statistical methods. Nonetheless, the noise sources vary dramatically on different sensors, making it impractical to extract and model all kinds of noise sources accurately. Zhang~\etal~\cite{zhang2021rethinking} proposes a novel strategy for synthesizing noisy raw images, which samples directly from a Poisson distribution and a database of dark frame images to obtain synthetic images. However, the dark frames vary with the exposure time and ISO level, which makes it challenging to obtain the dark frame database for all exposure times and ISO levels.

\textbf{DNN-based Noise Model.} Early works ~\cite{chen2018image, kim2019grdn} attempt to synthesize realistic noisy images by utilizing Generative Adversarial Networks (GANs), but they provide limited improvement in real-world denoising performance. Abdelhamed~\etal~\cite{abdelhamed2019noise} proposes a novel generative noise model, Noise-Flow, based on normalization flow to synthesize realistic noisy raw image, which is the first generative model that provides powerful noise modeling capabilities. Chang~\etal~\cite{chang2020learning} presents a camera-aware generative noise model termed CA-NoiseGAN to generate multiple noises for different camera sensors. Monakhova~\etal~\cite{monakhova2022dancing} introduces a physics-inspired generative noise model to learn the noise distribution of the camera sensor in low-light conditions, but this approach is limited to synthesizing noise that mimics a single ISO level, which is cumbersome in practical applications. In this work, we introduce a generative noise model to synthesize noise for different ISO levels and generalize it to various camera sensors.

\section{Methodology}
\label{method}
\subsection{Sensor Noise Formation}
To model more realistic low-light raw noise, it is necessary to understand the noise formation of Complementary Metal-Oxide-Semiconductor (CMOS)~\cite{cmosmarket}. For a raw image produced by a CMOS sensor, the raw image formation process from incident photons to digital values can be modeled as follows:
\begin{equation}
    D = KI+N\label{eq:image formation} ,
\end{equation}
\noindent where $K$ presents the gain of the whole system, $I$ stands for the number of photoelectrons stimulated by the scene radiation, $N$ denotes the summation of all noise sources physically caused by light or camera.

Due to several physical limitations of CMOS, noise occurs from various sources. According to the relation between noise and light intensity~\cite{healey1994radiometric,el2005cmos, farrell2008sensor}, we analyze the raw image formation in Eq.~\eqref{eq:image formation} and categorize the raw image noise into signal-dependent and signal-independent components:
\begin{equation}
    D = K(I+N_{dep})+N_{indep}\label{eq:noise formation} ,
\end{equation}
\noindent where $N_{dep}$ represents the signal-dependent noise, and $N_{indep}$ represents the signal-independent noise. Generally, the signal-dependent noise includes photon-shot noise and photo response non-uniformity. For the signal-independent noise, it includes read-out noise, fix pattern noise~\cite{snoeij2006cmos}, dark current noise~\cite{baer2006model}, quantization noise, flicker noise~\cite{barnes1966statistical}, thermal noise, reset noise~\cite{konnik2014high}, and so on.

\begin{figure*}[htbp]
\centering
\includegraphics[width=\textwidth]{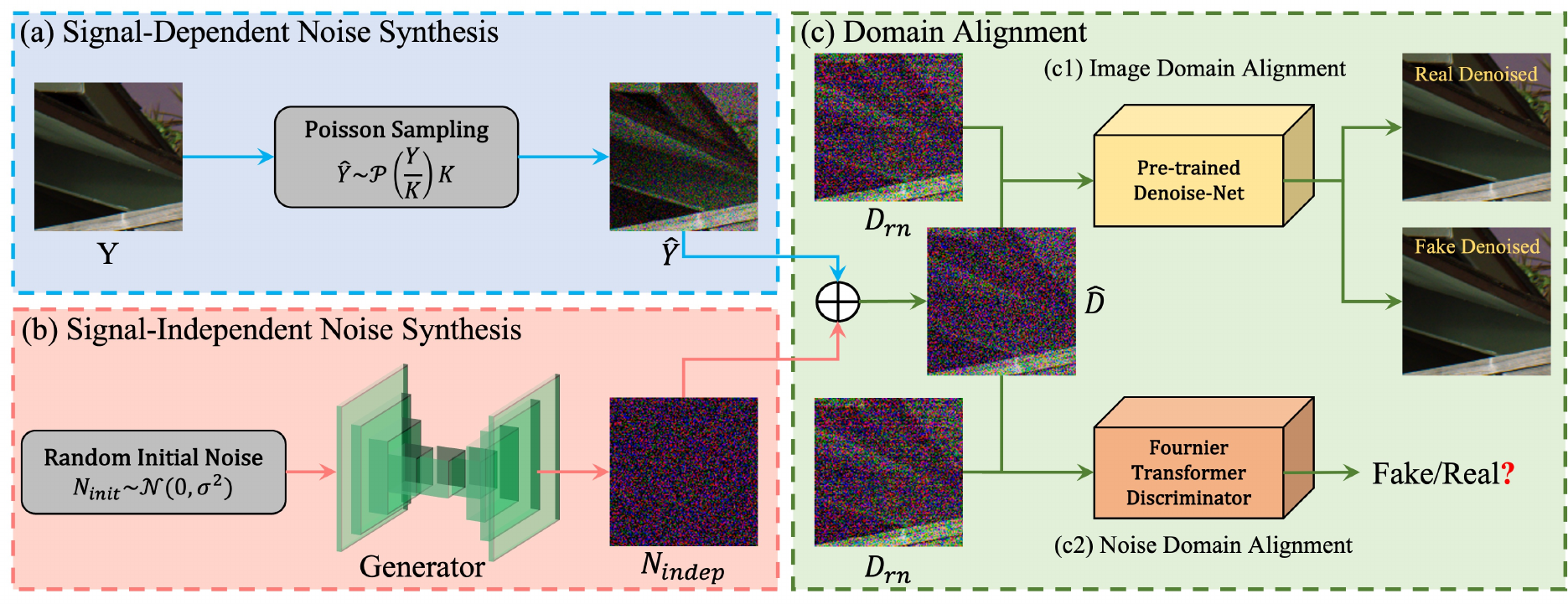}
\caption{Overview of the framework. The proposed noise model is divided into three components: (a) signal-dependent noise synthesis, (b) signal-independent noise synthesis, and (c) domain alignment. Please refer to Sec.~\ref{method} for more detailed descriptions.}
\label{fig arc}
\vspace{-2ex}
\end{figure*}

\subsection{Separated Noise Modeling}
According to the noise formation process described in Eq.\eqref{eq:noise formation}, we can separate the noise synthesis process into signal-dependent and signal-independent components. In well-lit environments, these two noise components can be accurately synthesized by physics-based models~\cite{zhang2021rethinking}. However, in low-light conditions, the signal-independent noise becomes more complex and varies significantly among camera sensors. Existing physics-based methods cannot accurately model this noise component, and building a noise model for each sensor is laborious. Therefore, we adopt a generative model to synthesize the signal-independent noise. For the signal-dependent noise, we still utilize a physics-based model for synthesis since most previous works~\cite{foi2009clipped,wei2020physics,zhang2021rethinking} have shown that the physics-based model can synthesize it accurately at a lower price. The framework of our proposed noise model is shown in Fig~\ref{fig arc}.

\textbf{Signal-Dependent Noise Synthesis.} From the noise formation process of CMOS sensors discussed above, we can observe that the signal-dependent noise consists of photon shot noise and photo response non-uniformity. However, according to previous reserches~\cite{gow2007comprehensive,janesick1987scientific}, the photo response non-uniformity contribute less than $3\%$ in the signal-dependent noise, which has a minimal impact. Therefore, the photon shot noise can be considered the only signal-dependent noise source.

Due to the intrinsic stochastic nature of photons reaching the CMOS sensor, the photon shot noise can be directly sampled from the Poisson distribution $\mathcal{P}$:
\begin{equation}
    (N_{dep}+I) \sim \mathcal{P}(I) .
\end{equation}

For the synthesis of signal-dependent noise, extensive studies~\cite{wei2020physics,zhang2021rethinking} have demonstrated that the incident photon numbers strictly follow the Poisson distribution. Thus we can accurately synthesize the signal-dependent noise in a physics-based manner, which can be modeled as follows:
\begin{equation}
    \hat{Y}=({N_{dep}}+I) = \mathcal{P}(\frac{Y}{K})K ,  
\end{equation}
\noindent where $Y$ is the clean image, $\hat{Y}$ is the sampled image contaminated by signal-dependent noise. The overall system gain $K$ can be easily obtained from the meta information of DNG file~\cite{adobe2012}.

\textbf{Signal-Independent Noise Synthesis.} Previous studies~\cite{wei2020physics, zhang2021rethinking} have demonstrated that signal-independent noise is the dominant component in low-light conditions. Since the noise sources in the signal-independent component are extraordinarily complicated and vary significantly with different exposure times, ISO levels, and camera sensors. Instead of adopting a physics-based approach to model the signal-independent noise, we exploit the powerful learning capabilities of the Generative Adversarial Networks (GAN) to model it. Fig.~\ref{fig arc} shows the overall framework of our proposed noise model.

To synthesize signal-dependent noise, we first sample a random initial noise map to reflect the stochastic noise behavior according to the conditions of each ISO level. Then, we fed the random initial noise map into a noise generator, which we utilize a typical residual 2D U-shape architecture~\cite{ronneberger2015u}. (More details of the generator architecture can be found in the supplementary material.) Formally, this synthesis process can be formulated as follow:
\begin{equation}
    N_{indep} = G(N_{init} \sim \mathcal{N}(0,\sigma_{r}^{2})) ,
\end{equation}
\noindent where $N_{init}$ is the sampled random initial noise map, $G$ is the proposed U-shape noise generator. We sample the random initial noise map from $\mathcal{N}(0,\sigma_{r}^{2})$ and spatially replicate through all pixel positions of the clean raw image $Y$. $\sigma_{r}$ is the noise parameter of the signal-independent component in the in-camera noise profile, which is related to the camera ISO levels. Similar to the conditional GAN~\cite{mirza2014conditional}, this parameter can be specified as a condition to control the noise level of $N_{init}$, which enables the network to generate different ISO levels of signal-independent noise.

Following the synthesis of signal-dependent and signal-independent noise by the two methods described above. Given a clean raw image $Y$, we can produce a pseudo-noisy raw image $\hat{D}$ as follows:
\begin{equation}\label{eq:generator}
    \hat{D} = K(I+N_{dep}) + G(N_{init}) = \hat{Y} + N_{indep} .
\end{equation}

In the training phase, we begin by taking paired raw images as inputs and subsequently extract essential parameters, including ISO levels, exposure times, and in-camera noise profiles from the noisy raw images, employing ExifTool. The generation of noisy raw images is accomplished by utilizing both the in-camera noise profile and the clean raw images as inputs to the generator.

During the inference phase, we adopt a different approach. Here, we leverage the in-camera noise profile of the target ISO level, in combination with the clean raw image, to serve as inputs. These inputs are then jointly fed into the generator, facilitating the synthesis of the noisy raw images corresponding to the desired target ISO level.

\textbf{Domain Alignment.} The most common strategy in image generation to construct image domain alignment is to minimize the distance between synthetic and real images directly. However, since the noise generator should produce different noise samples at each forward pass, it is incompatible with deploying the $\mathcal{L}_{1}$ loss directly between the synthesized noisy raw image $\hat{D}$ and the real noisy raw image $D_{rn}$. Besides, this strategy may drastically damage the quality of $\hat{D}$ due to the stochastic characteristics of noise~\cite{cai2021learning}. Therefore, the interference of noise should be excluded in the process of constructing image domain alignment. 

To tackle this issue, inspired by~\cite{cai2021learning}, we introduce a pre-trained denoise network~\cite{chen2018learning} (Denoise-Net) to transfer $\hat{D}$ and $D_{rn}$ into virtually noise-free image space. Then, perform $\mathcal{L}_{1}$ loss between the denoised versions of $\hat{D}$ and $D_{rn}$. Since the denoised image is relatively stable, minimizing the $\mathcal{L}_{1}$ loss enables the synthesized noise to converge to the real noise distribution while preserving the stochastic characteristics of the noise. The proposed $\mathcal{L}_{1}$ loss can be formulated as follows:
\begin{equation}
    \mathcal{L}_{1} = \parallel P(\hat{D}) - P(D_{rn}) \parallel_1 ,
\end{equation}
\noindent where $P$ prensent the Denoise-Net. Note that we utilize the same paired data to train the Denoise-Net and the noise model, thus eliminating potential domain gap issues.

We adopt adversarial learning for noise domain alignment to make the generated noise distribution fit real noise distribution as closely as possible. General discriminators perform superiorly in discriminating noise-free or noisy images with low noise levels. However, we find that they are insufficient to discriminate noisy raw images with a high noise level, especially for low-light noisy raw images. (See Table~\ref{table:ablation1}.) To tackle this issue, we introduce a new discriminator. See more details in the following subsection.

\subsection{Fourier Transformer Discriminator}
\label{ftd}
Based on the spectral transform theory, it is postulated that noise can be categorized as high-frequency information, whereas the content information of an image is typically associated with low-frequency components. Consequently, the operation performed in the spectral domain can more accurately differentiate between the noise and content information. Therefore, we present a transformer block called Fourier transformer block inspired by fast Fourier convolution (FFC)~\cite{chi2020fast}. The structure of this block is constructed by imitating the FFC, which replaces the convolution layer with the transformer block. (See supplementary material for the detailed structure.) Like the FFC, our proposed Fourier transformer block also takes two interlinked paths: a spectral path that conducts operation in the spectral domain with half of the input sequences and a spatial path that operates with another half. Each path receives complementary information and then performs an exchange internally to fuse the information. 

\begin{figure}[t]
\centering
\includegraphics[width=\linewidth]{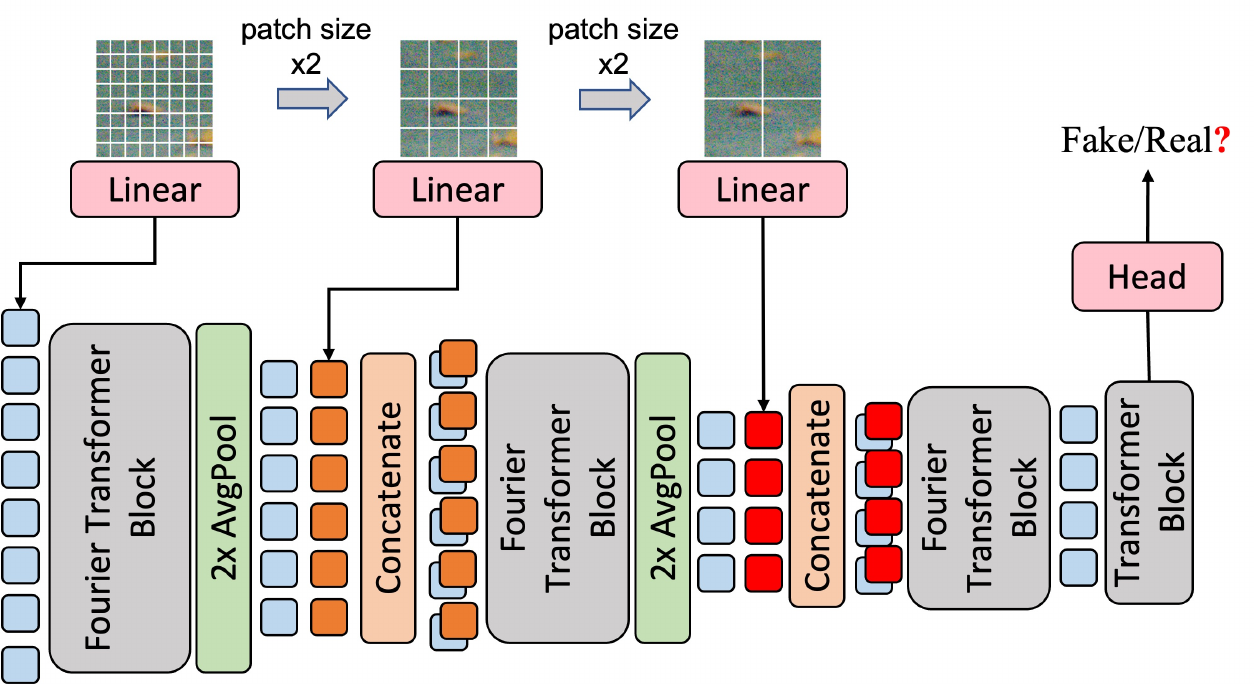}
\caption{Framework of the proposed Fourier Transformer Discriminator. The architecture comprises three Fourier transformer blocks and one vanilla transformer block.}
\label{fig:ffcd}
\vspace{-1ex}
\end{figure}

Motivated by TransGAN~\cite{jiang2021transgan}, we establish an effective multi-scale discriminator, namely Fourier transformer discriminator (FTD), based on the proposed Fourier transformer block. The overall framework of FTD is shown in Fig~\ref{fig:ffcd}. The 4-channel input raw image is firstly divided into three sequences by choosing different patch sizes. Then, these three sequences are fed into different Fourier transformer blocks through a linear transformation. The most extended sequence is combined with the learnable position encoding to serve as the input of the first block. The second and third sequences are concatenated into the second and third blocks. To reduce the resolution of the feature map between each block, we convert the 1D sequences to 2D feature maps and feed them into the Average Pooling layer. At last, a vanilla transformer block and a classification head are applied to output the prediction score.

\subsection{Training Objective}
To achieve the training objective, we optimize the generator and discriminator in an adversarial way~\cite{goodfellow2020generative}. Among adversarial frameworks, we select WGAN-GP~\cite{gulrajani2017improved} to calculate the adversarial loss $\mathcal{L}_{adv}$, which minimizes the Wasserstein distance to stabilize the training. The loss can be defined as follows:
\begin{multline}
    \mathcal{L}_{adv} = \underset{\hat{D} \sim \mathbb{P}_g}{\mathbb{E}} [{D}_{F}(\hat{D})] - \underset{{D}_{rn} \sim \mathbb{P}_r}{\mathbb{E}} [{D}_{F}({D}_{rn})]\\
    + \lambda \underset{\tilde{x} \sim \mathbb{P}_{\tilde{x}}}{\mathbb{E}} \| (\nabla_{\tilde{x}} {D}_{F}(\tilde{x})\|_2 - 1)^2] ,
\end{multline}
\noindent where ${D}_{F}$ is our proposed discriminator FTD, $\mathbb{P}_{g}$ is the synthetic noisy data distribution defined by the generator, $\mathbb{P}_{r}$ is the real noisy data distribution, ${D}_{rn}$ is the real noisy raw image.

In addition to the aforementioned $\mathcal{L}_{1}$ loss, we employ the perceptual loss to improve the perceptual quality:
\begin{equation}
\mathcal{L}_{per} = \parallel \phi(P(\hat{D}) - \phi(P(D_{rn})\parallel_2^2 ,
\end{equation}
\noindent where $\phi(\cdot)$ denotes the feature map extracted from a VGG-19~\cite{simonyan2014very} model pre-trained on ImageNet~\cite{deng2009imagenet}.

To summarize, the full objective of our proposed noise model is combined as follows:
\begin{equation}
\mathcal{L} = \mathcal{L}_{adv} + \lambda_{1}\mathcal{L}_{1} + \lambda_{2}\mathcal{L}_{per} ,
\end{equation}
\noindent where $\lambda_{1}$ and $\lambda_{2}$ are two hyper-parameters to control the balance of loss functions. In our experiment, these parameters are set to $\lambda_{1}= 0.1$, $\lambda_{2}= 0.01$.

\begin{figure}[t]
\centering
\includegraphics[width=\linewidth]{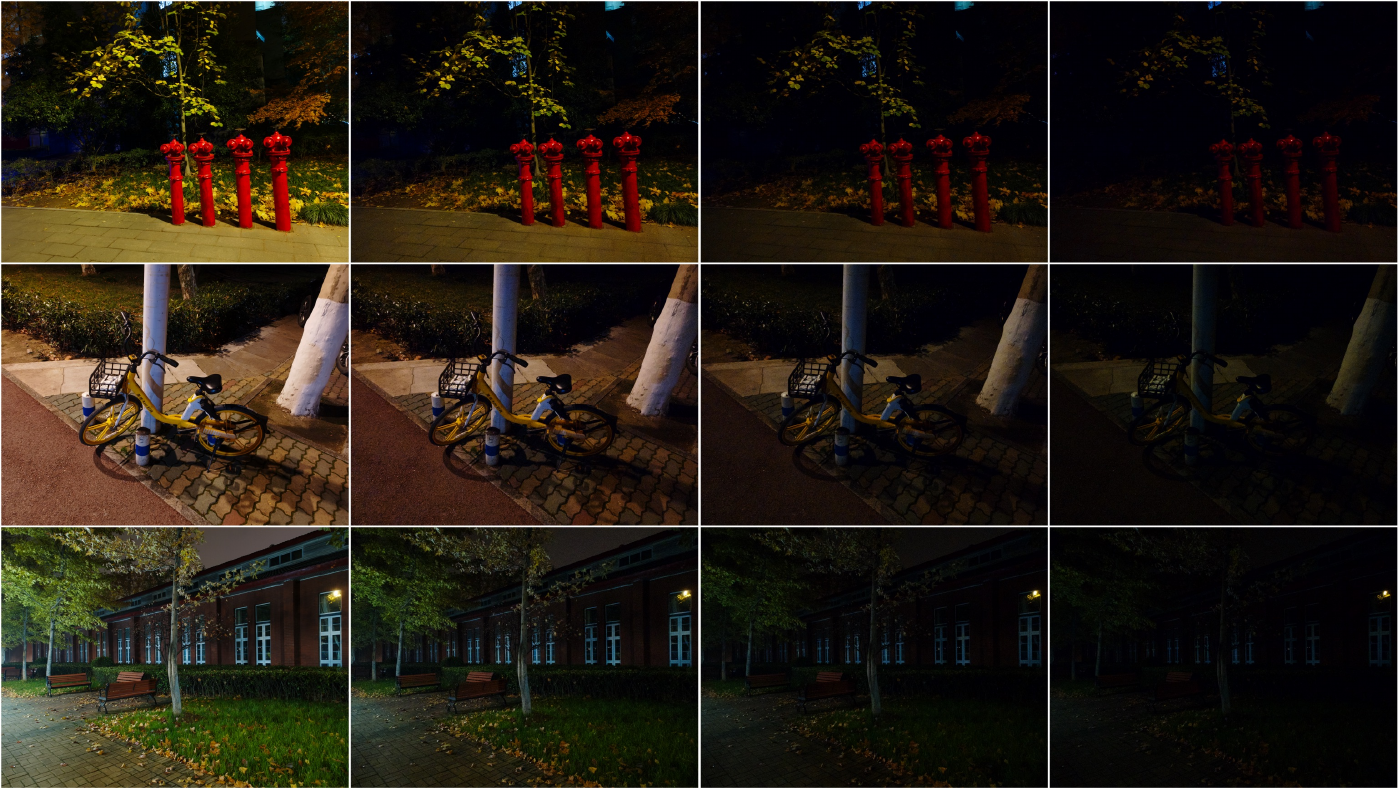}
\caption{Example images of the LRD dataset. First column: long exposure reference (ground truth) images. Second column: low-light images with -1EV. Third column: low-light images with -2EV. Fourth column: low-light images with -3EV.}
\label{fig:dataset}
\vspace{-1ex}
\end{figure}

\begin{table*}[!t]
\centering
\caption{Quantitative comparisons on the SID, ELD and LRD datasets in terms of PSNR and SSIM. The results are conducted on different exposure ratios. The \textcolor{red}{red} color indicates the best results, and the \textcolor{blue}{blue} color indicates the second-best results.}
\vspace{2pt}
\begin{tabular}{ccccccc}
\toprule[1.5pt]
\multirow{3}{*}{Dataset} & \multirow{3}{*}{Ratio} & \multicolumn{2}{c}{Physics-based}                 & \multicolumn{1}{c}{Real-noise-based} & \multicolumn{2}{c}{DNN-based}   \\ \cline{3-7} 
&  & \multicolumn{1}{c}{Poisson-Gaussian} & \multicolumn{1}{c}{ELD} & \multicolumn{1}{c}{Paired data} & \multicolumn{1}{c}{Noise Flow} & \multicolumn{1}{c}{Ours} \\ \cline{3-7} 
&  & PSNR / SSIM & PSNR / SSIM &PSNR / SSIM  & PSNR / SSIM & PSNR / SSIM\\\hline
\multirow{3}{*}{SID}& $\times100$ &37.51 / 0.856 &41.21 / 0.952 &\textcolor{blue}{41.39} / \textcolor{blue}{0.954} &36.75 / 0.787 &\textcolor{red}{41.95} / \textcolor{red}{0.956}\\
& $\times250$ &31.67 / 0.765 &38.54 / 0.929 &\textcolor{blue}{38.90} / \textcolor{red}{0.937} &33.98 / 0.739 & \textcolor{red}{39.25} / \textcolor{blue}{0.931}\\
& $\times300$ &28.53 / 0.667 &35.35 / 0.908 &\textcolor{red}{36.55} / \textcolor{red}{0.922} &31.82 / 0.713 &\textcolor{blue}{36.03} / \textcolor{blue}{0.909}\\ \hline
\multirow{2}{*}{ELD} &$\times100$ &39.46 / 0.785 &\textcolor{red}{45.06} / \textcolor{blue}{0.975} &43.80 / 0.963 &38.68 / 0.793 &\textcolor{blue}{44.95} / \textcolor{red}{0.979} \\
& $\times200$ &33.81 / 0.615 &\textcolor{blue}{43.21} / \textcolor{blue}{0.954} &41.54 / 0.918 &36.30 / 0.713 &\textcolor{red}{43.32} / \textcolor{red}{0.966}\\ \hline
\multirow{3}{*}{LRD}& -1EV &33.77 / 0.895 &38.31 / 0.968 &\textcolor{blue}{38.80} / \textcolor{blue}{0.970} & 35.19 / 0.874 & \textcolor{red}{38.89} / \textcolor{red}{0.971} \\
& -2EV &32.99 / 0.856 &37.35 / 0.959 &\textcolor{blue}{37.88} / \textcolor{blue}{0.961} &34.55 / 0.842 &  \textcolor{red}{37.95} / \textcolor{red}{0.962} \\
& -3EV &31.44 / 0.770 &36.49 / 0.950 &\textcolor{blue}{36.92} / \textcolor{blue}{0.951} &33.72 / 0.826 & \textcolor{red}{37.01} / \textcolor{red}{0.953} \\
\bottomrule[1.5pt]
\end{tabular}
\label{table:psnr}
\vspace{-2ex}
\end{table*}

\section{Low-light Raw Denoising (LRD) Dataset}
Creating low-light raw image datasets is essential for standardizing low-light raw denoising techniques. The existing dataset for benchmarking low-light raw denoising is the See-in-the-Dark (SID) dataset~\cite{chen2018learning}. However, since it is designed to produce perceptually good images in low-light conditions, there are some limitations in benchmarking low-light raw denoising. First, the long-exposure reference images in this dataset still contain some noise, which may disorient the generative noise model. Second, the varying number of images for each ISO level and exposure time may cause class imbalance issues. All these limitations hinder applications for low-light raw denoising and low-light raw image synthesis.

To address this issue, we collected a new low-light raw denoising (LRD) dataset for training and benchmarking. In contrast to the SID dataset, which sets a fixed exposure time to capture long and short exposure images, we captured long and short exposure images based on the exposure value (EV). Motivated by multi-exposure image fusion~\cite{ma2015perceptual, ma2019deep}, the exposure value for long exposure images was set to 0, and the exposure value for short exposure was set to the commonly used parameters -1, -2, and -3. The dataset is designed for application to low-light raw image denoising and low-light raw image synthesis.

The dataset contains both indoor and outdoor scenes. For each scene instance, we first captured a long-exposure image at ISO 100 to get a noise-free reference image. Then we captured multiple short-exposure images using different ISO levels and EVs, with a 1-2 second interval between subsequent images to wait for the sensor to cool down, thus avoiding unexpected noise introduced by sensor heating.

We captured 6 different ISO levels ranging from 200 to 6400 to obtain various noise levels. We captured 100 images at each ISO and EV setting to preserve a balanced training sample. Therefore, the total number of images in our dataset is 1800 (100 images $\times$ 6 ISO levels $\times$ 3 exposure value). Images were captured using a typical camera sensor: IMX586, which has a full-frame Bayer filter. The image resolution is $4000\times3000$. We mounted the camera on a sturdy tripod to ensure the sensor would not wobble. An example of long-short image pairs at different exposure values is shown in Fig.~\ref{fig:dataset}. We selected $10\%$ of the images in each exposure value and ISO levels to form the test set and selected another $5\%$ of the images as the validation set.


\begin{table}[!t]
\centering
\caption{Average Kullback-Leibler divergence (AKLD)~\cite{yue2020dual} evaluation of different noise models. Our proposed noise model outperforms state-of-the-art methods on both SID and LRD datasets.}
\vspace{2pt}
\scalebox{0.98}{
\begin{tabular}{cccccc}
\toprule[1.5pt]
Dataset &Ratio &P-G &ELD &Noise Flow &Ours \\ \hline
\multirow{3}{*}{SID} &$\times100$ &0.179 &0.117 &0.162 &\textbf{0.075}\\
 &$\times250$ &0.254 &0.177 &0.249 &\textbf{0.113}\\
 &$\times300$ &0.325 &0.231 &0.293 &\textbf{0.119}\\ \hline
 \multirow{3}{*}{LRD} &-1EV &0.459 &0.135 &0.274 &\textbf{0.099}\\
 &-2EV &0.592 &0.163 &0.338 &\textbf{0.109}\\
 &-3EV &0.701 &0.188 &0.375 &\textbf{0.123}\\
\bottomrule[1.5pt]
\end{tabular}}
\label{table:KLD}
\vspace{-3ex}
\end{table}

\vspace{-0.5em}
\section{Experiments}
\label{experiment}
\vspace{-0.5em}
\subsection{Experimental Setup}
\vspace{-0.5em}
\textbf{Dataset.} We first utilize the SID Sony set~\cite{chen2018learning} or our proposed LRD dataset to train the pre-trained denoise network. Then we fix the pre-trained denoise network to train the generator and discriminator on the same set. Subsequently, the generator takes the clean raw images from the SID Sony set or LRD dataset as input images to generate realistic pseudo-noisy raw images. The generated noisy-clean raw image pairs are utilized to evaluate the denoising performance on three benchmarks: the Sony set of SID and ELD datasets, and the LRD dataset. The images in the SID Sony set are collected using Sony $\alpha$7S\uppercase\expandafter{\romannumeral 2} in 231 static scenes. There are 1865 images for training, 234 for validation, and 598 for testing. The ELD Sony set~\cite{wei2020physics} consists of 60 low-light noisy raw images for testing, which are also captured using the same camera as the SID.

\begin{figure*}
\begin{center}
\includegraphics[width=\linewidth]{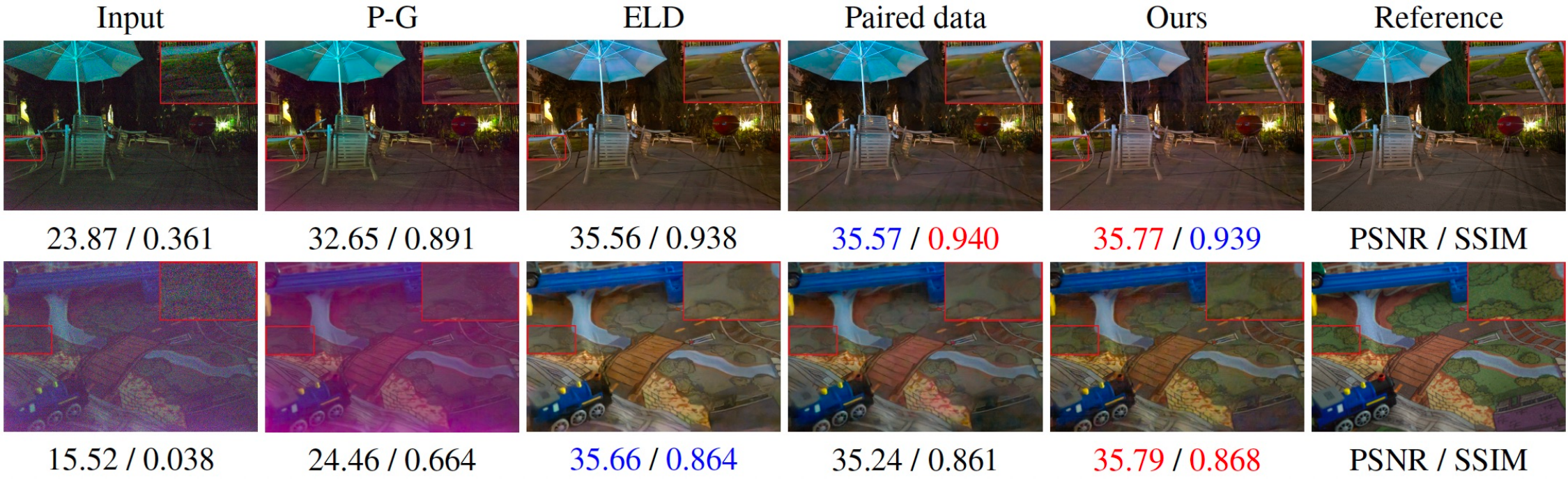}
\caption{Raw image denoising comparison with state-of-the-art methods on low-light noisy raw images from the SID dataset~\cite{chen2018learning}. Best viewed in color and by zooming in.}
\label{fig:SID}
\end{center}
\vspace{-3ex}
\end{figure*}

\begin{figure*}
\begin{center}
\includegraphics[width=\linewidth]{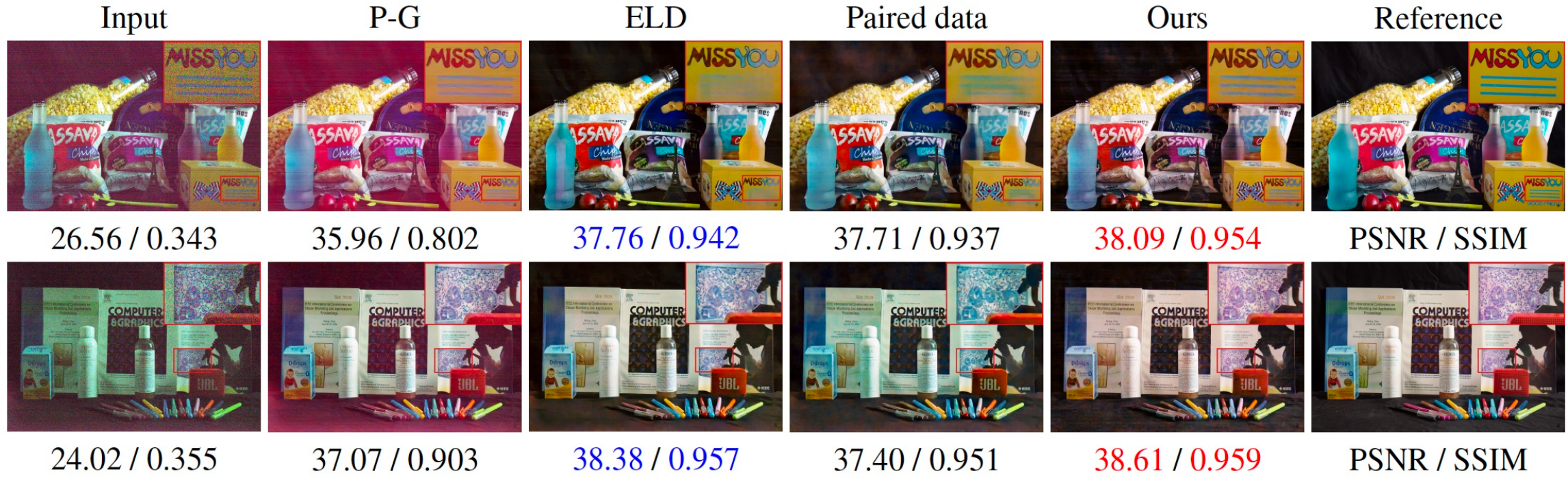}
\caption{Raw image denoising comparison with state-of-the-art methods on low-light noisy raw images from the ELD dataset~\cite{wei2020physics}. Best viewed in color
and by zooming in.}
\label{fig:ELD}
\end{center}
\vspace{-4ex}
\end{figure*}

\textbf{Implementation Details.} To optimize the noise generator and discriminator, we augment all training samples by randomly cropping and horizontal flipping to construct a mini-batch of size 128. The Adam~\cite{kingma2014adam} optimizer is adopted for training, while the initial parameters $\beta_{1}$ and $\beta_{2}$ to 0.5 and 0.999. The initial learning rate is set to $2\times10^{-4}$, and the patch size is set to $64\times64$. The cosine annealing strategy is employed to steadily decrease the learning rate from the initial value to $10^{-6}$ during the training procedure, where the model is trained over 100 epochs. Denoising models are optimized using the generated training pairs from the trained generator with a mini-batch of size 1. The patch size is set to $512\times512$. The Adam~\cite{kingma2014adam} optimizer is utilized with an initial learning rate of $2\times10^{-4}$, followed by halving at epoch 100 and finally to $5\times10^{-5}$ at epoch 180. The training runs for 200 epochs. We select the same U-shape architecture in~\cite{chen2018learning} as our denoising baseline. All of our experiments are conducted on four-channel raw images. The implementation is conducted on the Pytorch framework with a single GeForce 2080Ti GTX GPU.

\vspace{-0.5em}
\subsection{Model Analysis on Synthetic Noise}
\vspace{-0.5em}
We first analyze the realism of the synthesized pseudo-noise raw images generated by our proposed generator. For quantitative comparison, we utilize the widely applied metric, Average Kullback-Leibler divergence (AKLD)~\cite{yue2020dual}, to measure the discrepancy between the real noise and the synthetic noise patches generated by different noise formation models in the SID dataset and LRD dataset. The results are depicted in Table.~\ref{table:KLD}. Our proposed noise model achieves the minimum AKLD on both datasets. These results demonstrate that the distribution of synthetic noise generated by our proposed noise model more closely matches the real noise distribution.

\begin{figure*}
\begin{center}
\includegraphics[width=\linewidth]{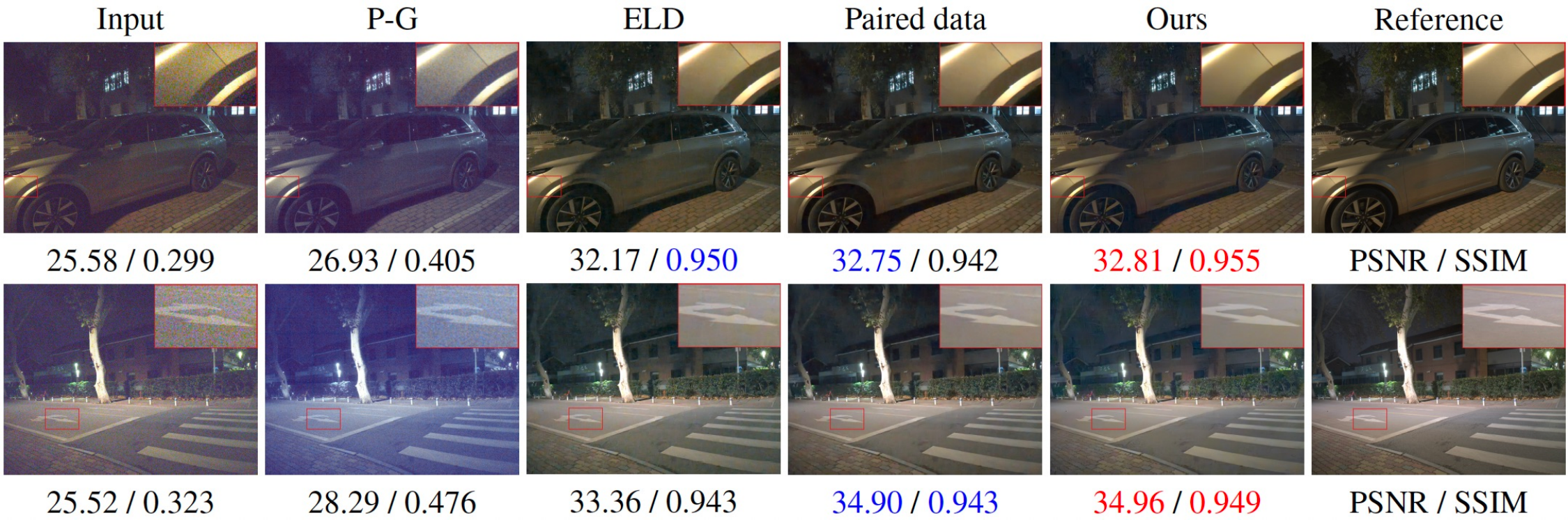}
\caption{Raw image denoising comparison with state-of-the-art methods on low-light noisy raw images from our proposed LRD dataset. Best viewed in color and by zooming in.}
\label{fig:LRD}
\end{center}
\vspace{-3ex}
\end{figure*}

\vspace{-0.5em}
\subsection{Denoising Results on SID and ELD datasets}
\vspace{-0.5em}
In order to demonstrate the reliability of our proposed noise model, we conduct raw image denoising experiments on SID and ELD datasets and compare them with some state-of-the-art methods: physics-based noise model Poisson-Gaussian (P-G), ELD, and DNN-based noise model Noise Flow. Moreover, we also compare our noise model with the model trained with paired real data.

Table.~\ref{table:psnr} shows the comparison results on the SID and ELD datasets over different exposure ratios. For the physics-based methods, the ELD noise model significantly outperforms the P-G noise model in terms of both PSNR and SSIM scores. As described in the previous work~\cite{zhang2021rethinking}, the DNN-based method Noise Flow performs poorly in low-light conditions, which indicates that this method is unsuitable for synthesizing low-light noisy raw images. Our noise model outperforms all existing low-light noise synthesis methods over different exposure ratios and even partially outperforms the denoiser trained with real paired data. This is because the real captured low-light raw image pairs still suffer from issues such as luminance misalignment and pixel misalignment~\cite{lehtinen2018noise2noise}, which may lead to unsatisfactory denoising results. A visual comparison of our noise model and other noise models is shown in Fig.~\ref{fig:SID} and Fig.~\ref{fig:ELD}. The P-G noise model is far from the real noise distribution. Thus it fails to remove the noise. The ELD noise model provides better denoising results but suffers from color bias and residual noise. The result of denoising with paired real data obtains favorable visual quality but still suffers from imperfections. Our proposed noise model can provide competitive denoising results with visual pleasurable.

\vspace{-0.5em}
\subsection{Denoising Results on LRD dataset}
\vspace{-0.5em}
To further validate the generalization ability of our proposed noise model, we assess model performance on our LRD dataset. Table.~\ref{table:psnr} and Fig.~\ref{fig:LRD} summarize all competing methods' quantitative and qualitative results. The results are consistent with the results on the SID and ELD datasets. Our method can still achieve comparable results with state-of-the-art methods. The P-G noise model still fails to remove the real noise, resulting in unsatisfactory visual results. ELD noise model provides satisfactory noise removal but still suffers from color bias issues. The results of our noise model are visually pleasing without significant residual noise or color bias, demonstrating our noise model's generalization ability across different camera devices.

\begin{table}[t]
\centering
\caption{Ablation study of the contribution of each component in the model framework in terms of PSNR and SSIM. The best results have been shown in bold.}
\vspace{2pt}
\begin{tabular}{c|c|c|c|c}
\hline
\multirow{2}{*}{\textbf{Components}} & \multirow{2}{*}{\textbf{Index}} & \multicolumn{3}{c}{\textbf{SID}} \\\cline{3-5}
& & \textbf{$\times100$} & \textbf{$\times250$} & \textbf{$\times300$} \\ \hline
\multirow{2}{*}{w/o Noise Separation} & PSNR &39.91 &37.53 &33.05 \\ \cline{2-5}
& SSIM &0.902 &0.883 &0.867  \\ \hline
\multirow{2}{*}{w/o Denoise-Net} & PSNR &41.09 &38.83 &35.66 \\ \cline{2-5}
& SSIM &0.937 &0.913 &0.899  \\ \hline
\multirow{2}{*}{Full framework} & PSNR &\textbf{41.95} &\textbf{39.25} &\textbf{36.03} \\ \cline{2-5}
& SSIM &\textbf{0.956} &\textbf{0.931} &\textbf{0.909}\\\hline
\end{tabular}
\label{table:ablation2}
\vspace{-2ex}
\end{table}

\vspace{-0.5em}
\subsection{Ablation Study}
\label{ablation}
\vspace{-0.5em}
\textbf{Impact of Each Components.} We perform break-down ablations to evaluate each component's effects in the model framework. The comparison results evaluated on the SID dataset are reported in Table.~\ref{table:ablation2}. First, we remove the noise separation strategy and directly synthesize the noise using the generative model, we follow the scheme of CA-NoiseGAN~\cite{chang2020learning} and take the noise parameter instead of the initial Gaussian noise map as the condition information. The PSNR and SSIM values show a significant decrease. Secondly, after deploying the pre-trained denoise network (Denoise-Net), the performance of the noise model is improved moderately, suggesting that the pre-trained denoising network successfully performs image domain alignment and improves the quality of synthetic images. These results convincingly demonstrate the superiority of our proposed noise model in synthesizing low-light raw noise.

\begin{table}[t]
\centering
\caption{Ablation study of the contribution of Fourier transformer discriminator in terms of PSNR, SSIM. The best results have been shown in bold.}
\vspace{2pt}
\begin{tabular}{c|c|c|c|c}
\hline
\multirow{2}{*}{\textbf{Discriminator}} & \multirow{2}{*}{\textbf{Index}} & \multicolumn{3}{c}{\textbf{SID}} \\\cline{3-5}
& & \textbf{$\times100$} & \textbf{$\times250$} & \textbf{$\times300$} \\ \hline
\multirow{2}{*}{Transformer} & PSNR &40.23 &37.36 &33.22 \\ \cline{2-5}
& SSIM &0.931 &0.910 &0.892\\\hline
\multirow{2}{*}{Fourier Transformer} & PSNR &\textbf{41.95} &\textbf{39.25} &\textbf{36.03} \\ \cline{2-5}
& SSIM &\textbf{0.956} &\textbf{0.931} &\textbf{0.909}\\\hline
\end{tabular}
\label{table:ablation1}
\vspace{-2ex}
\end{table}

\textbf{Effectiveness of Fourier Transformer Discriminator.} A significant advantage of our proposed noise model over existing generative noise models is that the FTD can effectively fuse information through operations in spectral and spatial domains. To verify the effectiveness of the FTD, we employ a conventional discriminator with vanilla transformer blocks for comparison. Table.~\ref{table:ablation1} shows the comparison results on the SID dataset, and our proposed FTD can discriminate the noise distribution more accurately, thus encouraging the generator to synthesize more accurate noise.

\section{Conclusion}
\label{conclusion}
In this paper, we present a new perspective for realistic low-light raw noise synthesis. Specifically, we synthesize the signal-dependent and signal-independent noise in a physics- and learning-based manner. We employ a pre-trained denoise network during the training procedure to transfer the synthesized and real noisy raw images into a nearly noise-free image space for image domain alignment. Meanwhile, we introduce an effective discriminator, namely Fourier transformer discriminator (FTD), to perform noise domain alignment. Our method is general for different ISO levels and different camera sensors. Subsequently, we collect a new low-light raw denoising (LRD) dataset for training and benchmarking. Both qualitative and quantitative experiments on the public datasets and our dataset collectively demonstrate the superiority of our method over state-of-the-art methods.

\textbf{Acknowledgements.} This work was supported by the National Natural Science Foundation of China No.62176097, Hubei Provincial Natural Science Foundation of China No.2022CFA055.

{\small
\bibliographystyle{ieee_fullname}
\bibliography{reference}

\begin{thebibliography}{10}\itemsep=-1pt

\bibitem{abdelhamed2019noise}
Abdelrahman Abdelhamed, Marcus~A Brubaker, and Michael~S Brown.
\newblock Noise flow: Noise modeling with conditional normalizing flows.
\newblock In {\em Proceedings of the IEEE/CVF International Conference on
  Computer Vision}, pages 3165--3173, 2019.

\bibitem{abdelhamed2018high}
Abdelrahman Abdelhamed, Stephen Lin, and Michael~S Brown.
\newblock A high-quality denoising dataset for smartphone cameras.
\newblock In {\em Proceedings of the IEEE Conference on Computer Vision and
  Pattern Recognition}, pages 1692--1700, 2018.

\bibitem{aharon2006k}
Michal Aharon, Michael Elad, and Alfred Bruckstein.
\newblock K-svd: An algorithm for designing overcomplete dictionaries for
  sparse representation.
\newblock {\em IEEE Transactions on signal processing}, 54(11):4311--4322,
  2006.

\bibitem{baer2006model}
Richard~L Baer.
\newblock A model for dark current characterization and simulation.
\newblock In {\em Sensors, Cameras, and Systems for Scientific/Industrial
  Applications VII}, volume 6068, pages 37--48. SPIE, 2006.

\bibitem{barnes1966statistical}
JA Barnes and DW Allan.
\newblock A statistical model of flicker noise.
\newblock {\em Proceedings of the IEEE}, 54(2):176--178, 1966.

\bibitem{brooks2019unprocessing}
Tim Brooks, Ben Mildenhall, Tianfan Xue, Jiawen Chen, Dillon Sharlet, and
  Jonathan~T Barron.
\newblock Unprocessing images for learned raw denoising.
\newblock In {\em Proceedings of the IEEE/CVF Conference on Computer Vision and
  Pattern Recognition}, pages 11036--11045, 2019.

\bibitem{buades2005non}
Antoni Buades, Bartomeu Coll, and J-M Morel.
\newblock A non-local algorithm for image denoising.
\newblock In {\em 2005 IEEE computer society conference on computer vision and
  pattern recognition (CVPR'05)}, volume~2, pages 60--65. Ieee, 2005.

\bibitem{cai2021learning}
Yuanhao Cai, Xiaowan Hu, Haoqian Wang, Yulun Zhang, Hanspeter Pfister, and
  Donglai Wei.
\newblock Learning to generate realistic noisy images via pixel-level
  noise-aware adversarial training.
\newblock {\em Advances in Neural Information Processing Systems},
  34:3259--3270, 2021.

\bibitem{chang2020learning}
Ke-Chi Chang, Ren Wang, Hung-Jin Lin, Yu-Lun Liu, Chia-Ping Chen, Yu-Lin Chang,
  and Hwann-Tzong Chen.
\newblock Learning camera-aware noise models.
\newblock In {\em European Conference on Computer Vision}, pages 343--358.
  Springer, 2020.

\bibitem{chen2018learning}
Chen Chen, Qifeng Chen, Jia Xu, and Vladlen Koltun.
\newblock Learning to see in the dark.
\newblock In {\em Proceedings of the IEEE conference on computer vision and
  pattern recognition}, pages 3291--3300, 2018.

\bibitem{chen2018deep}
Chang Chen, Zhiwei Xiong, Xinmei Tian, and Feng Wu.
\newblock Deep boosting for image denoising.
\newblock In {\em Proceedings of the European Conference on Computer Vision
  (ECCV)}, pages 3--18, 2018.

\bibitem{chen2018image}
Jingwen Chen, Jiawei Chen, Hongyang Chao, and Ming Yang.
\newblock Image blind denoising with generative adversarial network based noise
  modeling.
\newblock In {\em Proceedings of the IEEE conference on computer vision and
  pattern recognition}, pages 3155--3164, 2018.

\bibitem{chen2015learning}
Yunjin Chen, Wei Yu, and Thomas Pock.
\newblock On learning optimized reaction diffusion processes for effective
  image restoration.
\newblock In {\em Proceedings of the IEEE conference on computer vision and
  pattern recognition}, pages 5261--5269, 2015.

\bibitem{chi2020fast}
Lu Chi, Borui Jiang, and Yadong Mu.
\newblock Fast fourier convolution.
\newblock {\em Advances in Neural Information Processing Systems},
  33:4479--4488, 2020.

\bibitem{dabov2006image}
Kostadin Dabov, Alessandro Foi, Vladimir Katkovnik, and Karen Egiazarian.
\newblock Image denoising with block-matching and 3d filtering.
\newblock In {\em Image processing: algorithms and systems, neural networks,
  and machine learning}, volume 6064, pages 354--365. SPIE, 2006.

\bibitem{dabov2007image}
Kostadin Dabov, Alessandro Foi, Vladimir Katkovnik, and Karen Egiazarian.
\newblock Image denoising by sparse 3-d transform-domain collaborative
  filtering.
\newblock {\em IEEE Transactions on image processing}, 16(8):2080--2095, 2007.

\bibitem{deng2009imagenet}
Jia Deng, Wei Dong, Richard Socher, Li-Jia Li, Kai Li, and Li Fei-Fei.
\newblock Imagenet: A large-scale hierarchical image database.
\newblock In {\em 2009 IEEE conference on computer vision and pattern
  recognition}, pages 248--255. Ieee, 2009.

\bibitem{el2005cmos}
Abbas El~Gamal and Helmy Eltoukhy.
\newblock Cmos image sensors.
\newblock {\em IEEE Circuits and Devices Magazine}, 21(3):6--20, 2005.

\bibitem{elad2006image}
Michael Elad and Michal Aharon.
\newblock Image denoising via sparse and redundant representations over learned
  dictionaries.
\newblock {\em IEEE Transactions on Image processing}, 15(12):3736--3745, 2006.

\bibitem{farrell2008sensor}
Joyce Farrell, Michael Okincha, and Manu Parmar.
\newblock Sensor calibration and simulation.
\newblock In {\em Digital Photography IV}, volume 6817, pages 249--257. SPIE,
  2008.

\bibitem{feng2022learnability}
Hansen Feng, Lizhi Wang, Yuzhi Wang, and Hua Huang.
\newblock Learnability enhancement for low-light raw denoising: Where paired
  real data meets noise modeling.
\newblock In {\em Proceedings of the 30th ACM International Conference on
  Multimedia}, pages 1436--1444, 2022.

\bibitem{foi2009clipped}
Alessandro Foi.
\newblock Clipped noisy images: Heteroskedastic modeling and practical
  denoising.
\newblock {\em Signal Processing}, 89(12):2609--2629, 2009.

\bibitem{foi2008practical}
Alessandro Foi, Mejdi Trimeche, Vladimir Katkovnik, and Karen Egiazarian.
\newblock Practical poissonian-gaussian noise modeling and fitting for
  single-image raw-data.
\newblock {\em IEEE Transactions on Image Processing}, 17(10):1737--1754, 2008.

\bibitem{gharbi2016deep}
Micha{\"e}l Gharbi, Gaurav Chaurasia, Sylvain Paris, and Fr{\'e}do Durand.
\newblock Deep joint demosaicking and denoising.
\newblock {\em ACM Transactions on Graphics (ToG)}, 35(6):1--12, 2016.

\bibitem{goodfellow2020generative}
Ian Goodfellow, Jean Pouget-Abadie, Mehdi Mirza, Bing Xu, David Warde-Farley,
  Sherjil Ozair, Aaron Courville, and Yoshua Bengio.
\newblock Generative adversarial networks.
\newblock {\em Communications of the ACM}, 63(11):139--144, 2020.

\bibitem{gow2007comprehensive}
Ryan~D Gow, David Renshaw, Keith Findlater, Lindsay Grant, Stuart~J McLeod,
  John Hart, and Robert~L Nicol.
\newblock A comprehensive tool for modeling cmos image-sensor-noise
  performance.
\newblock {\em IEEE Transactions on Electron Devices}, 54(6):1321--1329, 2007.

\bibitem{gu2014weighted}
Shuhang Gu, Lei Zhang, Wangmeng Zuo, and Xiangchu Feng.
\newblock Weighted nuclear norm minimization with application to image
  denoising.
\newblock In {\em Proceedings of the IEEE conference on computer vision and
  pattern recognition}, pages 2862--2869, 2014.

\bibitem{gulrajani2017improved}
Ishaan Gulrajani, Faruk Ahmed, Martin Arjovsky, Vincent Dumoulin, and Aaron~C
  Courville.
\newblock Improved training of wasserstein gans.
\newblock {\em Advances in neural information processing systems}, 30, 2017.

\bibitem{guo2019toward}
Shi Guo, Zifei Yan, Kai Zhang, Wangmeng Zuo, and Lei Zhang.
\newblock Toward convolutional blind denoising of real photographs.
\newblock In {\em Proceedings of the IEEE/CVF conference on computer vision and
  pattern recognition}, pages 1712--1722, 2019.

\bibitem{hasinoff2014photon}
Samuel~W Hasinoff.
\newblock Photon, poisson noise.
\newblock {\em Computer Vision, A Reference Guide}, 4, 2014.

\bibitem{healey1994radiometric}
Glenn~E Healey and Raghava Kondepudy.
\newblock Radiometric ccd camera calibration and noise estimation.
\newblock {\em IEEE Transactions on Pattern Analysis and Machine Intelligence},
  16(3):267--276, 1994.

\bibitem{adobe2012}
ADOBE~SYSTEMS INCORPORATED.
\newblock Digital negative (dng) specification.
\newblock \url {
  https://www.adobe.com/content/dam/acom/en/products/photoshop/pdfs/dng_spec_1.4.0.0.pdf
  }, 2012.

\bibitem{janesick1987scientific}
James~R Janesick, Tom Elliott, Stewart Collins, Morley~M Blouke, and Jack
  Freeman.
\newblock Scientific charge-coupled devices.
\newblock {\em Optical Engineering}, 26(8):692--714, 1987.

\bibitem{jiang2021transgan}
Yifan Jiang, Shiyu Chang, and Zhangyang Wang.
\newblock Transgan: Two pure transformers can make one strong gan, and that can
  scale up.
\newblock {\em Advances in Neural Information Processing Systems},
  34:14745--14758, 2021.

\bibitem{kim2019grdn}
Dong-Wook Kim, Jae Ryun~Chung, and Seung-Won Jung.
\newblock Grdn: Grouped residual dense network for real image denoising and
  gan-based real-world noise modeling.
\newblock In {\em Proceedings of the IEEE/CVF Conference on Computer Vision and
  Pattern Recognition Workshops}, pages 0--0, 2019.

\bibitem{kingma2014adam}
Diederik~P Kingma and Jimmy Ba.
\newblock Adam: A method for stochastic optimization.
\newblock {\em arXiv preprint arXiv:1412.6980}, 2014.

\bibitem{konnik2014high}
Mikhail Konnik and James Welsh.
\newblock High-level numerical simulations of noise in ccd and cmos
  photosensors: review and tutorial.
\newblock {\em arXiv preprint arXiv:1412.4031}, 2014.

\bibitem{lehtinen2018noise2noise}
Jaakko Lehtinen, Jacob Munkberg, Jon Hasselgren, Samuli Laine, Tero Karras,
  Miika Aittala, and Timo Aila.
\newblock Noise2noise: Learning image restoration without clean data.
\newblock In {\em International Conference on Machine Learning}, pages
  2965--2974. PMLR, 2018.

\bibitem{ma2019deep}
Kede Ma, Zhengfang Duanmu, Hanwei Zhu, Yuming Fang, and Zhou Wang.
\newblock Deep guided learning for fast multi-exposure image fusion.
\newblock {\em IEEE Transactions on Image Processing}, 29:2808--2819, 2019.

\bibitem{ma2015perceptual}
Kede Ma, Kai Zeng, and Zhou Wang.
\newblock Perceptual quality assessment for multi-exposure image fusion.
\newblock {\em IEEE Transactions on Image Processing}, 24(11):3345--3356, 2015.

\bibitem{maggioni2012video}
Matteo Maggioni, Giacomo Boracchi, Alessandro Foi, and Karen Egiazarian.
\newblock Video denoising, deblocking, and enhancement through separable 4-d
  nonlocal spatiotemporal transforms.
\newblock {\em IEEE Transactions on image processing}, 21(9):3952--3966, 2012.

\bibitem{mildenhall2018burst}
Ben Mildenhall, Jonathan~T Barron, Jiawen Chen, Dillon Sharlet, Ren Ng, and
  Robert Carroll.
\newblock Burst denoising with kernel prediction networks.
\newblock In {\em Proceedings of the IEEE conference on computer vision and
  pattern recognition}, pages 2502--2510, 2018.

\bibitem{mirza2014conditional}
Mehdi Mirza and Simon Osindero.
\newblock Conditional generative adversarial nets.
\newblock {\em arXiv preprint arXiv:1411.1784}, 2014.

\bibitem{monakhova2022dancing}
Kristina Monakhova, Stephan~R Richter, Laura Waller, and Vladlen Koltun.
\newblock Dancing under the stars: video denoising in starlight.
\newblock In {\em Proceedings of the IEEE/CVF Conference on Computer Vision and
  Pattern Recognition}, pages 16241--16251, 2022.

\bibitem{plotz2017benchmarking}
Tobias Plotz and Stefan Roth.
\newblock Benchmarking denoising algorithms with real photographs.
\newblock In {\em Proceedings of the IEEE conference on computer vision and
  pattern recognition}, pages 1586--1595, 2017.

\bibitem{portilla2003image}
Javier Portilla, Vasily Strela, Martin~J Wainwright, and Eero~P Simoncelli.
\newblock Image denoising using scale mixtures of gaussians in the wavelet
  domain.
\newblock {\em IEEE Transactions on Image processing}, 12(11):1338--1351, 2003.

\bibitem{cmosmarket}
Grand~View Research.
\newblock Image sensors market analysis.
\newblock
  \url{http://www.grandviewresearch.com/industry-analysis/imagesensors-market},
  2016.

\bibitem{ronneberger2015u}
Olaf Ronneberger, Philipp Fischer, and Thomas Brox.
\newblock U-net: Convolutional networks for biomedical image segmentation.
\newblock In {\em Medical Image Computing and Computer-Assisted
  Intervention--MICCAI 2015: 18th International Conference, Munich, Germany,
  October 5-9, 2015, Proceedings, Part III 18}, pages 234--241. Springer, 2015.

\bibitem{rudin1992nonlinear}
Leonid~I Rudin, Stanley Osher, and Emad Fatemi.
\newblock Nonlinear total variation based noise removal algorithms.
\newblock {\em Physica D: nonlinear phenomena}, 60(1-4):259--268, 1992.

\bibitem{simonyan2014very}
Karen Simonyan and Andrew Zisserman.
\newblock Very deep convolutional networks for large-scale image recognition.
\newblock {\em arXiv preprint arXiv:1409.1556}, 2014.

\bibitem{snoeij2006cmos}
Martijn~F Snoeij, Albert~JP Theuwissen, Kofi~AA Makinwa, and Johan~H Huijsing.
\newblock A cmos imager with column-level adc using dynamic column
  fixed-pattern noise reduction.
\newblock {\em IEEE Journal of Solid-State Circuits}, 41(12):3007--3015, 2006.

\bibitem{wang2019enhancing}
Wei Wang, Xin Chen, Cheng Yang, Xiang Li, Xuemei Hu, and Tao Yue.
\newblock Enhancing low light videos by exploring high sensitivity camera
  noise.
\newblock In {\em Proceedings of the IEEE/CVF International Conference on
  Computer Vision}, pages 4111--4119, 2019.

\bibitem{wang2020practical}
Yuzhi Wang, Haibin Huang, Qin Xu, Jiaming Liu, Yiqun Liu, and Jue Wang.
\newblock Practical deep raw image denoising on mobile devices.
\newblock In {\em European Conference on Computer Vision}, pages 1--16.
  Springer, 2020.

\bibitem{wei2020physics}
Kaixuan Wei, Ying Fu, Jiaolong Yang, and Hua Huang.
\newblock A physics-based noise formation model for extreme low-light raw
  denoising.
\newblock In {\em Proceedings of the IEEE/CVF Conference on Computer Vision and
  Pattern Recognition}, pages 2758--2767, 2020.

\bibitem{yue2019variational}
Zongsheng Yue, Hongwei Yong, Qian Zhao, Deyu Meng, and Lei Zhang.
\newblock Variational denoising network: Toward blind noise modeling and
  removal.
\newblock {\em Advances in neural information processing systems}, 32, 2019.

\bibitem{yue2020dual}
Zongsheng Yue, Qian Zhao, Lei Zhang, and Deyu Meng.
\newblock Dual adversarial network: Toward real-world noise removal and noise
  generation.
\newblock In {\em Computer Vision--ECCV 2020: 16th European Conference,
  Glasgow, UK, August 23--28, 2020, Proceedings, Part X 16}, pages 41--58.
  Springer, 2020.

\bibitem{zhang2017beyond}
Kai Zhang, Wangmeng Zuo, Yunjin Chen, Deyu Meng, and Lei Zhang.
\newblock Beyond a gaussian denoiser: Residual learning of deep cnn for image
  denoising.
\newblock {\em IEEE transactions on image processing}, 26(7):3142--3155, 2017.

\bibitem{zhang2021rethinking}
Yi Zhang, Hongwei Qin, Xiaogang Wang, and Hongsheng Li.
\newblock Rethinking noise synthesis and modeling in raw denoising.
\newblock In {\em Proceedings of the IEEE/CVF International Conference on
  Computer Vision}, pages 4593--4601, 2021.

\end{thebibliography}
}

\end{document}